\documentclass[letterpaper, 10 pt, conference]{ieeeconf}  % Comment this line out if you need a4paper

\IEEEoverridecommandlockouts                              % This command is only needed if 
\title{\LARGE \bf
Privacy-Preserving Pose Estimation for Human-Robot Interaction
}

\author{Youya Xia$^{1^{*}}$, Yifan Tang$^{1^{*}}$, Yuhan Hu$^{1}$ and Guy Hoffman$^{1}$% <-this % stops a space
\thanks{*These authors contribute equally to the work.}% <-this % stops a space
\thanks{$^{1}$Youya Xia, Yifan Tang, Yuhan Hu and Guy Hoffman are with
        Cornell University
        }%
}

\usepackage{amsmath}
\usepackage{array}
\usepackage{url} 
\usepackage{caption}
\usepackage{subcaption}
\usepackage{enumerate}
\usepackage{adjustbox}
\usepackage{paralist}
\usepackage{longtable}
\usepackage{multirow} % tables
\usepackage{rotating} % for rotating texts
\usepackage{float}
\usepackage{url}
\usepackage{paralist}
\usepackage{amsfonts}
\usepackage{longtable}
\usepackage{graphicx}

\usepackage{times} % assumes new font selection scheme installed
\usepackage{amsmath} % assumes amsmath package installed
\usepackage{amssymb}  % assumes amsmath package installed
\usepackage{amsfonts}
\usepackage{nicefrac}
\usepackage{algorithm} %[boxed]
\usepackage{algpseudocode}
\usepackage{wrapfig}

\usepackage{url}
\usepackage[compress]{cite}
\usepackage{paralist}

\graphicspath{{images/}}
\usepackage{ifpdf}
\ifpdf\DeclareGraphicsExtensions{.pdf,.png,.jpg}\fi

\newcommand{\bma}{\begin{bmatrix}}
\newcommand{\ema}{\end{bmatrix}}

\long\def\comment#1{}

\usepackage[dvipsnames]{xcolor}
\usepackage[utf8x]{inputenc}

\makeatletter
\let\NAT@parse\undefined
\makeatother
\begin{document}

\maketitle
\thispagestyle{empty}
\pagestyle{empty}

%%%%%%%%%%%%%%%%%%%%%%%%%%%%%%%%%%%%%%%%%%%%%%%%%%%%%%%%%%%%%%%%%%%%%%%%%%%%%%%%
%%%%%%%%%%%%
\begin{abstract}
   Pose estimation is an important technique for nonverbal human-robot interaction. That said, the presence of a camera in a person's space raises privacy concerns and could lead to distrust of the robot. In this paper, we propose a privacy-preserving camera-based pose estimation method. The proposed system consists of a user-controlled translucent filter that covers the camera and an image enhancement module designed to facilitate pose estimation from the filtered (shadow) images, while never capturing clear images of the user. We evaluate the system's performance on a new filtered image dataset, considering the effects of distance from the camera, background clutter, and film thickness. Based on our findings, we conclude that our system can protect humans' privacy while detecting humans' pose information effectively. 
\end{abstract}
\section{Introduction}

% - Simple summary
In this paper, we describe a method for privacy-preserving visual human-robot interaction (HRI). 
Our approach rests on effective human pose inference from a user-degraded camera.
This would allow a user to cover the robot's camera with a translucent film but still interact with the robot using nonverbal behaviors. 

% - Motivation
Visual observation of humans plays a vital role in understanding humans' states and intentions for HRI~\cite{goodrich2008human}. However, being monitored by cameras, especially at home, leads to privacy concerns~\cite{caine2012effect}. %Thus, estimating humans' poses while still protecting humans' privacy is essential for robots to interact with humans fluently.
%%Prior literature for privacy protection 
Prior work on protecting humans' privacy in captured images focuses on using depth cameras or specially designed low-resolution cameras~\cite{srivastav2019human, lyu2017privacy}. However, these methods require specialized equipment. %Therefore, they lack feasibility and cannot be configured easily. 
In contrast, we propose to improve pose inference in images captured from a standard RGB camera which has been covered by a translucent film. The main contribution of this work is an image enhancement method that maintains privacy but enables robust pose estimation. 

%%prior literature for solving pose estimation problems on degraded images

Pose estimation on clear images has been studied extensively~\cite{herath2017going, sarafianos20163d}, but conducting pose estimation directly on degraded images is challenging. Researchers have proposed a number of image enhancement methods to facilitate pose estimation for degraded images~\cite{li2017aod, li2020underwater}. However, these methods are generally based on a specific image formation model, such as low light or haze. We found that the privacy-preserving images in our case do not comply with existing models. Therefore, the existing models cannot solve the pose estimation problems for our purpose (see Section~\ref{subsec:compare_contrast} for a detailed evaluation).

\begin{figure}[t]
    \centering
    \includegraphics[width=\linewidth]{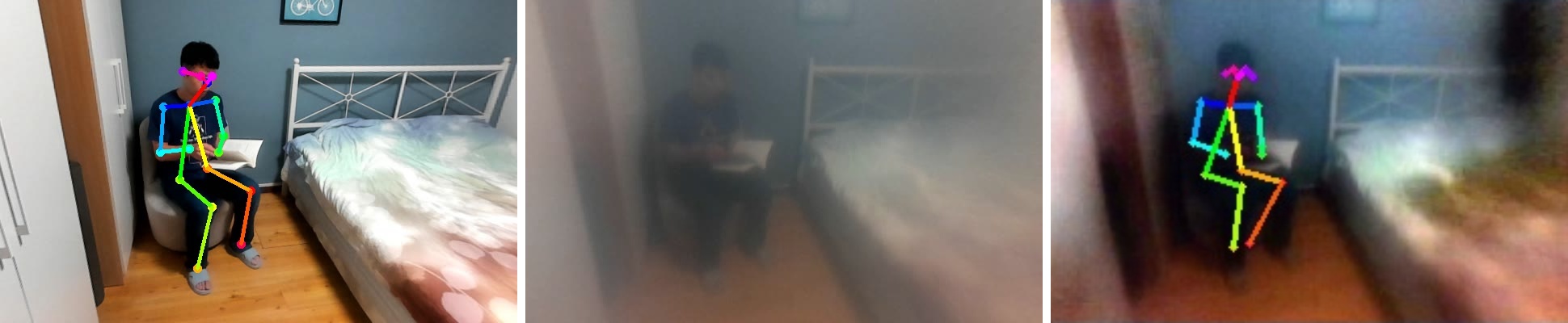}
    \caption{Left to right: the pose estimation results on the ground-truth clear image, the corresponding shadow image and enhanced image output by our enhancement module.}
    \label{fig:intro_fig}
\end{figure}

Our method is building on the recommendation put forth in Hu \textit{et al.}~\cite{shadowsense}, which proposed privacy-preserving HRI with cameras covered up by filtering materials. 
To detect humans' poses from the degraded images, we propose a neural network based architecture for image enhancement. The proposed network is trained to produce an enhanced version of the input image. Afterwards, the enhanced image is passed into \textit{OpenPose}~\cite{cao2019openpose} to obtain the final estimated poses (Figure~\ref{fig:intro_fig}).
Experimental results indicate that our system can obtain human pose information effectively. %Besides, we offer future designers suggestions about choosing suitable setups (\textit{e.g.}, filters, proximity) for the privacy-preserving camera system.

The paper thus makes the following contributions:
\begin{itemize}
    \item A design for camera-based human-robot interaction, which can protect humans' privacy while obtaining their pose information effectively;
    \item A neural network architecture designed for image enhancement specifically aimed at pose estimation from translucent-film-filtered images\footnote{The training and testing code is released at: \small\url{https://github.com/xiaxx244/shadow_pose_estimation}}; 
    \item An extensive quantitative evaluation of the proposed system providing insights on the effects of distance, background clutter, and film thickness.
\end{itemize}

\section{Related Work}
In this section, we review prior works on privacy-enhanced HRI and on image enhancement for pose estimation.

\subsection{Privacy Protection in HRI}

The camera used in many robots as an input to monitor a user's state or intentions carries with it privacy concerns, especially in personal spaces such as the home. Researchers have proposed privacy-protecting strategies, such as blur filtration of all or part of the image~\cite{fan2005novel, neustaedter2006blur}. As these methods aim to protect humans' privacy through processing already-taken video data, they still allow robots to capture high-fidelity visual data.

Users often take control of their camera privacy by physically obstructing the lens of a camera placed in a private space. 
Building on this practice, Hu \textit{et al.} propose a privacy-maintaining possibility for camera-based interaction with social robots~\cite{shadowsense}. The authors suggest that through physically covering the robot's cameras with a translucent material, a robot can still exploit some interaction data in the form of users' shadows instead of high-fidelity images.
The paper, however, only described a concept sketch of this idea, without proposing algorithms for inferring users' poses from full-body shadow images. 
Establishing an effective design for such an algorithm is the aim of the current work, and at its core lies an appropriate image enhancement module. 

%%Yuhan TODO here, some hint and high-level writing instructions below:
% \iffalse
% We will review the traditional methodologies for privacy protection in HRI
% Current privacy concerns in human-robot interaction and proposed approaches to address/protect; especially when interacting with a camera-mounted robot.
% Also a short review of what benefits/information can come with when using robot vision.
% The importance of estimating human activities for human-robot interaction with some examples.(2-3 convincing examples)
% \fi
 
\subsection{Image Enhancement for Pose Estimation}
Image enhancement is an active field of computer vision, with existing methods broadly falling into two categories: single-image dehazing and low-light image enhancement. 

\paragraph{Single-image Dehazing}
Haze is a common type of image degradation, leading to two kinds of work related to single-image dehazing. The first line of work focuses on estimating transmission maps and atmospheric light. It then uses this atmospheric scattering model to remove haze. For example, Li \textit{et al.} (2017) provides an end-to-end network that directly estimates the combination of atmospheric light and transmission map~\cite{li2017aod}. The other line of work focuses on learning a
mapping from a hazy image to a haze-free image. For instance, Li \textit{et al.} (2020) proposes a convolutional neural network (CNN) to learn a residual between hazy underwater images and haze-free images~\cite{li2020underwater}. But, both lines of work do not have robust performances when images contain non-uniform light or heavy haze~\cite{Ren2016,ren2020single}.

\paragraph{Low-Light Image Enhancement}
Image generated from low-light circumstances is another type of image degradation. The early literature about low-light image enhancement focuses on contrast enhancement~\cite{pisano1998contrast, cheng2004simple, abdullah2007dynamic, celik2011contextual}. However, the early literature's enhanced images suffer from problems such as image distortion and artificial illumination. State-of-the-art methodologies~\cite{wang2013naturalness,fu2016fusion,lv2018mbllen,fu2016weighted} have improved the naturalness of illumination and the realism of images significantly. Nevertheless, they still encounter failures of enhancement in regions with complex textures.

Both lines of state-of-the-art image enhancement methodologies focus on degraded images with explicit formation models. Since there is no such formation model for images generated with a film-covered camera, implementing the state-of-the-art methodologies in our method is not a promising avenue. In Section~\ref{sec:experiments} we explicitly evaluate the use of existing image enhancement methods in comparison with the method propose in this paper. 

\section{System}
This section describes the full privacy-preserving pose estimation pipeline, which includes the privacy-preserving camera setup, the proposed network for image enhancement module, and the pose estimation module.

%%In ths section, Yuhan TODO:
%%describe why we choose to use filters to cover cameras as modified camera-based interaction
\subsection{Privacy-preserving camera setups}
Based on the prior work of ShadowSense~\cite{shadowsense}, in the proposed system, the robots' eyes are covered with a translucent material which allows it to capture users' full-body visual data in the form of their shadows. This is a solution that can easily be implemented by the end-user and leaves them in full control of the image captured by the robot.

While adding privacy, conducting pose estimation directly on the defined shadow images is challenging (see Section~\ref{sec:experiments} for detailed performance evaluation of pose estimation on these images). To counter this issue, we present an image enhancement module that can produce an enhanced version of a shadow image tailored to human pose estimation.

\begin{figure}[h]
\centering
%\hspace{-3mm}
\includegraphics[width=0.7\linewidth, height=1.9cm]{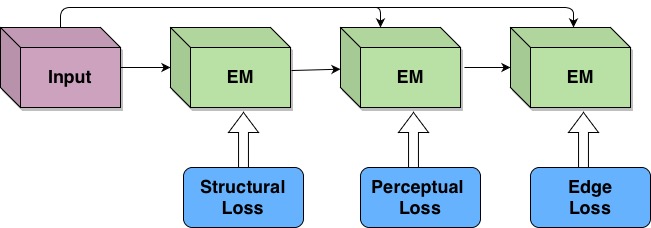}
%\vspace{-10mm}
\caption{The general structure of the \textit{MiniRes}-based architecture. It is composed of an input layer with size $256 \times 256 \times 3$ and three Enhancement Modules (EM), each of which has a respective optimization objective (\textit{i.e.}, structural loss, perceptual loss and edge loss). The second and the third modules are connected with the input layer via a \textit{short-cut} respectively.}
%\vspace{-8mm}
\label{fig:arch}
\end{figure} 
\begin{figure}[h]
\centering
%\hspace{-3mm}
\includegraphics[width=\linewidth]{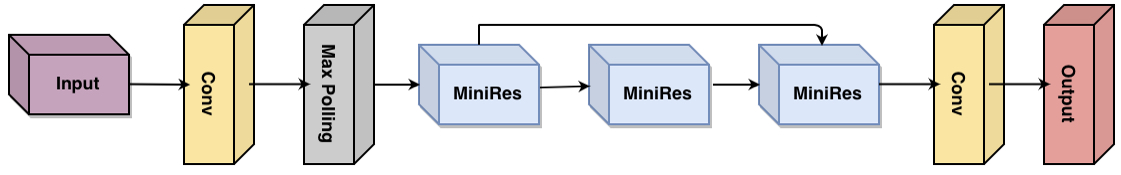}
%\vspace{-10mm}
\caption{The architecture of each Enhancement Module (EM). It consists of 3 \textit{MiniRes} blocks with a \textit{short-cut} from the first \textit{MiniRes} block to the third \textit{MiniRes} block.}
%\vspace{-8mm}
\label{fig:EM}
\end{figure} 
\iffalse
\begin{figure}[t]
\centering
%\hspace{-3mm}
\includegraphics[width=0.4\linewidth]{Figs/minires.jpg}
%\vspace{-10mm}
\caption{The architecture of MiniRes block. we modify the Resblock~\cite{he2016deep} to remove the $2$ batch normalization layers. To achieve faster inference, each of the convolution layers have a dimension of $3 \times 3 \times 32$.}
%\vspace{-8mm}
\label{fig:minires}
\end{figure} 
\fi

\subsection{\textit{MiniRes}-based architecture for Image Enhancement}
% The overall structure of the network is explained in this paragraph

We propose a design for deep neural network (Figure~\ref{fig:arch}), composed of an input layer with size $256 \times 256 \times 3$ and three separate ``Enhancement Modules'' (EM).
Each EM is tuned to a specific optimization objective which is important to pose estimation.

The first enhancement module aims to restore the shadow image's structural information so that the enhanced image has higher structural similarity to the ground-truth clear image. In the second enhancement module, the optimization objective is to restore the shadow image's perceptual features so that the enhanced image has a smaller distance to the ground-truth image in visual space. The final enhancement module is supposed to restore the edge information of the shadow image so that the enhanced image preserves the edge information of the ground-truth image. The three EMs are connected in sequential order, and the output of our network is the output from the third enhancement module. Moreover, to avoid classical degradation and vanishing gradient problems encountered by deep neural networks, we add a shortcut from the input layer to the second and the third enhancement modules, respectively. 

% The details of the enhancement modules
Inside each of the enhancement modules (Figure~\ref{fig:EM}), the input is connected to a 3D Convolution layer whose dimension is $3 \times 3 \times 32$ and a MaxPooling layer whose filter size is $3 \times 3$.
We then use three sequential copies of a custom-designed component which is based on the \textit{ResBlock} structure in \textit{ResNet}~\cite{he2016deep}, an efficient network structure for various challenging vision tasks. We call this comoponent \textit{MiniRes}.

\textit{MiniRes} blocks are different from original \textit{ResBlocks} in two ways. First, to allow for the variance of shadow images created by different filters, we remove the two batch normalization layers from the original \textit{ResBlock} architecture. Second, to speed up the inference process, we reduce the dimension of two 3D convolution layers from $3 \times 3 \times 64$ to $3 \times 3 \times 32$.  

In addition, in each EM, a shortcut is added from the first to the third \textit{MiniRes} Block to preserve information about the image's content during optimization process. Finally, we connect the third \textit{MiniRes} block to a 3D convolution layer with dimension $3 \times 3 \times 32$ to produce the output of size $256 \times 256 \times 3$ for each enhancement module.

\subsection{Loss Function Formulation}
Each EM uses a module-specific loss function, with the total loss function defined as
\begin{equation}
    \mathcal{L}=\mathcal{L}_{SL}+\mathcal{L}_{PL}+\mathcal{L}_{EL}
\end{equation}
Here, $\mathcal{L}_{SL}$ represents the structural loss, $\mathcal{L}_{PL}$ represents the perceptual loss and $\mathcal{L}_{EL}$ represents the edge loss, mapping to the three optimization objectives listed above. 

\begin{itemize}
%%Structural Loss modified here after technical proofreading paper details;
\item The \textbf{structural loss $\mathcal{L}_{SL}$} aims to restore the structural information of the shadow image's content. To do so, we use the \textit{Structural Similarity Index Measure} (SSIM)~\cite{wang2004image}, which has shown success for repairing an image's structural degradation~\cite{lv2018mbllen, li2019heavy}. For each pixel $x$, SSIM is calculated within a window size $11\times 11$ around $x$ as follows: 
    \begin{equation}
\text{SSIM}(x):=\frac{(2\mu_{e}(x)\mu_{c}(x)+d_{1})(2\sigma_{e*c}(x)+d_{2})}{(\mu_{e}^{2}(x)+\mu_{c}^{2}(x)+d_{1})(\sigma_{e}^{2}(x)+\sigma_{c}^{2}(x)+d_{2})}
    \end{equation}
Here, $e$ and $c$ correspond to the network's latent image and the paired ground-truth clear image.
$\mu, \sigma^{2}$  correspond to expectation and variance of pixel values lying in each image window, and $\sigma_{e*c}^{2}$ corresponds to the the pixels' covariance between paired windows in $e$ and $c$. $d_{1}$ and $d_{2}$ are regulating constants which are set to $0.0001$ and $0.0009$ respectively. Given this definition, $\mathcal{L}_{SL}$ is defined as:
\begin{equation}
    \mathcal{L}_{SL}:=1-\frac{1}{3M}\sum_{i=1}^{M}(\text{SSIM}_{r}(e_{i})+\text{SSIM}_{g}(e_{i})+\text{SSIM}_{b}(e_{i}))
\end{equation}
where $M$ is the number of pixels in the network's latent image $e$ and $\text{SSIM}_{h}(e_{i})$ means for each image channel $h \in \{r,g,b\}$, the SSIM value at pixel $i$ inside $e$.
   
\item The \textbf{perceptual Loss} $\mathcal{L}_{PL}$ ensures that the network's output image is similar to the clear image in visual space.
\begin{equation}
    \mathcal{L}_{PL}:=\mathcal{L}_\text{MSE}+2\mathcal{L}_\text{MAE}+\mathcal{L}_\textit{ResNet}
\end{equation}
    We use a mixture of the Mean Absolute Error (MAE) (the $L_{1}$ norm) and the Mean Squared Error (MSE) (the $L_{2}$ norm), as the mixture has been proved to be a more robust metric for comparing global similarity in pixel space than using either norm alone~\cite{fu2005fast}.
    In addition, because \textit{ResNet} has already been shown to be a robust feature extractor for high-level vision tasks, we use a feature extractor $\phi$ based on the top 6 layers of  \textit{ResNet-50}~\cite{he2016deep} pretrained on \textit{ImageNet}~\cite{deng2009imagenet}. Then we compute the $L_{2}$ distance between the network's latent image $e$ and the
    ground-truth image $c$ in the \textit{ResNet} feature space: 
    \begin{equation}
        \mathcal{L}_\textit{ResNet}:= ||\phi(e)-\phi(c)||_{2}
    \end{equation}
      
    \item Finally, the \textbf{edge loss} $\mathcal{L}_{EL}$ aims to restore the edge information of the ground-truth image. Specifically, we use an edge map extractor $\Omega$ which extracts the Sobel Edge~\cite{schalkoff1989digital} maps of the network's latent image $e$ and the corresponding ground-truth image $c$. Then, we computed the $L_{2}$ distance of those extracted maps as follows: 
    \begin{equation}
         \mathcal{L}_{EL}:=||\Omega(e)-\Omega(c)||_{2}
    \end{equation}
    
\end{itemize}

\subsection{Network Training}
%%Yifan TODO: describe briefly the metrics you choose to evaluate
There exists no large-scale image dataset with film-filtered camera images. Based on prior literature of image enhancement~\cite{li2017aod, li2020underwater}, training only on the limited number of shadow images that we would be able to collect does not produce robust performances. We therefore chose to train the proposed image enhancement module on the Outdoot Training Set (OTS)~\cite{li2018benchmarking}. The OTS  is a large-scale synthetic dataset consisting of degraded images with different degrees of degradation. We train our proposed \textit{MiniRes}-based network on the OTS using two NVIDIA\textsuperscript{TM} RTX 2080 Ti cards.

\subsection{Pose Estimation Module}
After the image enhancement module, the proposed system uses a pose estimation module to produce the final estimated poses in the enhanced image. Although various solutions for human body-pose detection have been proposed such as \textit{FastPose}~\cite{shakhnarovich2003fast} and other skeleton trackers~\cite{kohli2013key}, \textit{OpenPose} is established as the state-of-the-art~\cite{cao2019openpose}. In the last part of the proposed pipeline, we use \textit{OpenPose}, as is, on the enhanced images as its inputs, to extract points lying inside humans' bodies. 
% After the image enhancement module, the proposed system uses a pose estimation module to produce the final estimated poses in the enhanced image. We use \textit{OpenPose} as the pose estimation module to extract points lying inside humans' bodies for enhanced images. The reason to choose openPose is that it is the most successful solutions compared to other human body-pose detection method such as FastPose detectors~\cite{shakhnarovich2003fast} and the Kinect skeleton trackers~\cite{kohli2013key}. 
\begin{figure}[h]
\centering
 \includegraphics[width=0.6\linewidth,height=1.9995cm]{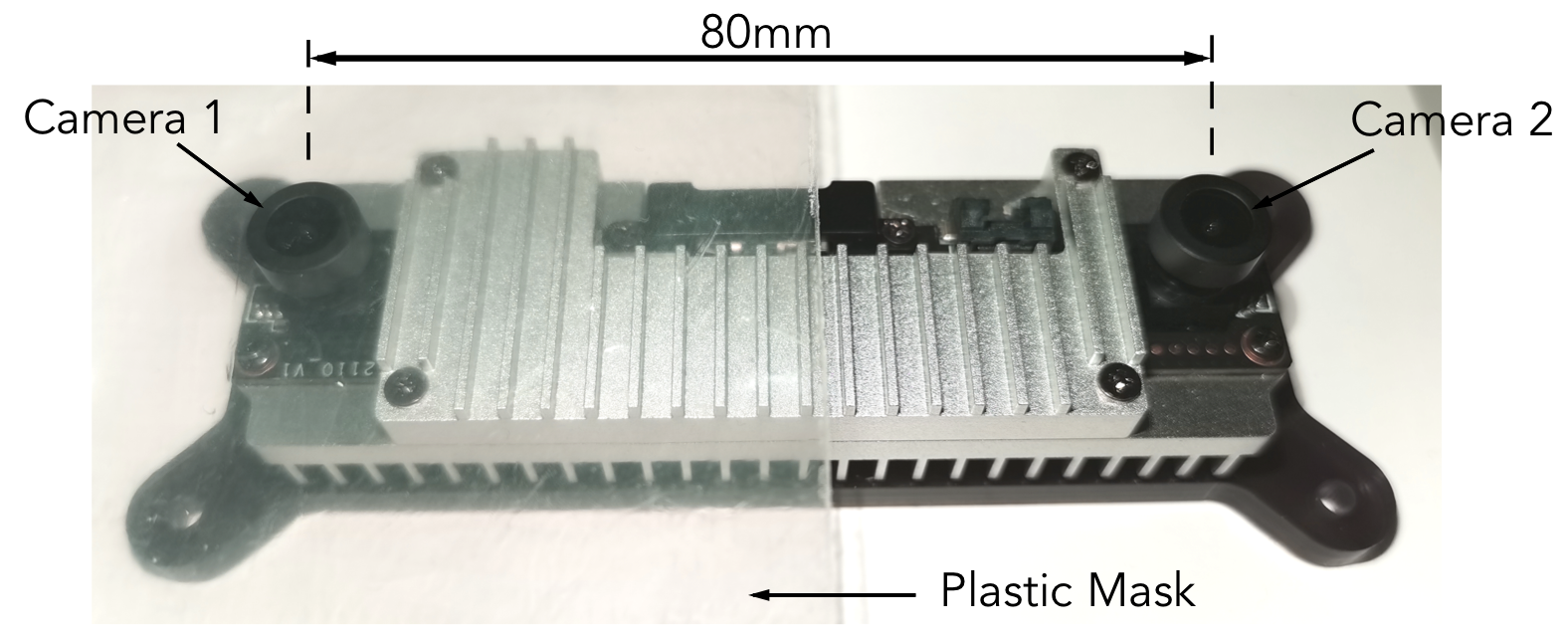}
  %\caption{The camera system with plastic bags on it.}
\caption{The figure shows the system when we put $1$ layer of the plastic mask (\textit{i.e.}, one of our chosen filter layers) on top of the left camera.}
\label{fig:camera}
\end{figure}
\section{Data Collection}
To  test our image enhancement method, we collected a ``shadow image'' dataset using a binocular camera system (Figure~\ref{fig:camera}), covering up one of the two cameras when capturing an image. Using a binocular camera ensured that the ground-truth image and the paired shadow image include the same content. To reduce the impact of binocular disparity~\cite{qian1997binocular}, we ensure that the distance between humans and the camera during the data collection process is at least $2$m.
\begin{figure}[ht]
  \vspace*{\fill}
  \centering
  \includegraphics[width=\linewidth]{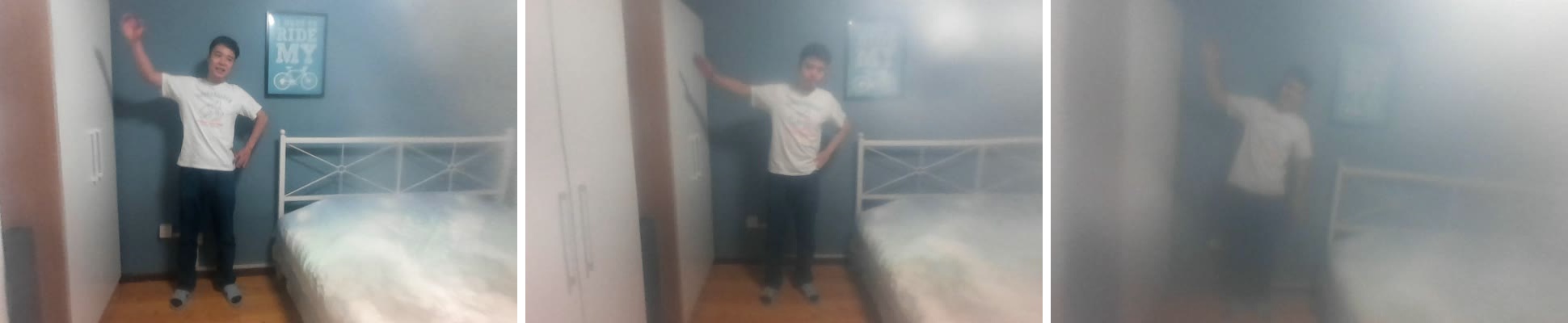}
  \caption{The shadow images created by our chosen filter layers. Left to right: shadow images created by $1$, $2$ and $3$ filter layers respectively.}
  \label{fig:layers}\par\vfill
\end{figure}

% Methodology to collect data. Justify the method: the distance between humans and the camera. 
Then, to create the shadow effect in one of the two cameras in the system, we used a freezer bag as the ``base'' filter. To evaluate the proposed system with different levels of privacy protection, we use $1$, $2$, \& $3$ layers of the freezer bag. Figure~\ref{fig:layers} shows the shadow effects created by selected layers. 
% The detail of the dataset
To test our proposed system's robustness, we conducted the experiments under different background complexities (\textit{i.e.}, plain backgrounds, and complex backgrounds). Environments containing less than 4 objects are defined as plain backgrounds, denoted ``private'' spaces below. Similarly, we define environments containing more than 3 objects as complex backgrounds and we call them ``public'' spaces. 
\iffalse
\begin{figure}[ht]
\centering
  \includegraphics[width=\linewidth]{Figs/act.png}
\caption{The sample ground-truth clear images show the activities used in our data collection process. Left to right images show the reading, exercising and drinking activities.}
\label{fig:activites}
\end{figure}
\fi
In addition, since \textit{OpenPose} uses a bottom-up approach for pose estimation, the farther distance between humans and cameras causes more challenges for the body part detection branch in its architecture. Aiming to understand the effectiveness of the \textit{MiniRes}-based network for enhancing details (\textit{i.e.}, smaller body parts under farther distance) in the shadow image, we also conduct experiments under different proximity (\textit{i.e.}, close \& far). Finally, to make sure there are no gender biases in our dataset, we collected images with both a female and a male actor. 

We asked the two actors to conduct 3 activities (reading, exercising, and drinking) in both public and private spaces. The actors were also requested to repeat the activities with different distances ($2$ \& $4$ meters) to the camera system.  In total, we have collected $72$ paired datasets in public and private spaces which are composed of $99,450$ paired images. In each type of spaces, we have collected $6$ paired datasets for each activity under close and far distances respectively.  
\section{Experiments and Metrics}
\label{sec:experiments}
This section conducts a quantitative evaluation of the proposed system, using the collected test shadow image dataset. We first evaluate the effectiveness of chosen filter layers, the robustness of the proposed image enhancement module quantitatively. Based on the proposed system's performances, we then offer future researchers potential directions about how to choose suitable filters and proximity for setting up a privacy-preserving camera system for a social robot.

\subsection{Filter Layers: Effectiveness Evaluation of Privacy Protection}
To provide a quantitative measure of the filter layers' effectiveness for privacy protection, we use an established measure of image quality and estimate the subsequent degradation.
Spatial–Spectral Entropy-based Quality (SSEQ)~\cite{liu2014no} is a comprehensive metric for measuring image quality based on images' visual and structural information without using ground-truth images as references. 

\begin{table}[h]
\begin{center}
\begin{tabular}{|c|c|c|c|}
\hline
        & $\text{SSEQ}_\text{Clear}$ & $\text{SSEQ}_\text{Shadow}$ & SR \\ \hline
1 Layer & 42.7710      & 52.6125      &  0.2301    \\ \hline
2 Layers & 42.7596     & 58.4768      &  0.3676     \\ \hline
3 Layers & 39.0658     & 57.7562      &   0.4784    \\ \hline
\end{tabular}
\end{center}
\caption{The SSEQ scores for our chosen filter layers in both private and public spaces. We also include the Shadow Ratio scores in the table to demonstrate the each filter layers' ability to protect humans' identities.}
\label{table:sseq}

\end{table}

Table~\ref{table:sseq} shows the mean SSEQ score for clear and shadow dataset, as well as a Shadow Ratio (SR) defined as 
\begin{equation}
    \text{SR}:=\frac{\text{SSEQ}_\text{Shadow}-\text{SSEQ}_\text{Clear}}{\text{SSEQ}_\text{Clear}}
    \end{equation}

Table~\ref{table:sseq} suggests that a single layer of film has the lowest image degradation, and this the lowest potential for privacy protection, whereas three layers of film has the strongest degradation and thus the most potential for privacy protection. This can also be seen qualitatively in the illustrative images in Figure~\ref{fig:layers}.

\subsection{Image Enhancement Metrics}
The proposed image enhancement module aims to facilitate the precision of pose estimation. So to evaluate the robustness of the trained \textit{MiniRes}-based network, we propose two evaluation metrics based on the pose estimation results we obtain from the \textit{OpenPose} module. 
\paragraph{Detection} The first metric we propose is Detection Rate, $\text{DR}:=\frac{N_{e}}{N_{c}}$. Here, $N_{e}$ stands for the detected skeleton key points in the shadow images or enhanced images, while $N_{c}$ stands for the detected key points in the paired ground-truth clear images. This metric only measures the number of detected features, without taking into account their accuracy.
\paragraph{Precision} Inspired by the mean Average Precision metric on pose estimation~\cite{andriluka20142d}, we also use a second metric, the Shadow mean Average Precision,
$ \text{SmAP}:=\frac{N_{te}}{N_{e}}$
Here, $N_{te}$ stands for the number of the precisely detected key points in the shadow or enhanced images. Specifically, for each detected keypoint $k$ in a shadow or enhanced image, we compare its location with the keypoint's location associated with the same body part in the paired clear image and calculate their $L_{2}$ distance as $L_{ec}$. To count for the precision errors arising from binocular disparity which, in our dataset, is around $10$ pixels, we regard $k \in N_{te}$ if $L_{ec} \leq 10$ pixels. 

We will compare both metrics across filter layers, distance from camera, and background clutter.

\section{Results}
We evaluate the robustness of the trained network on the test dataset collected in public and private spaces using the proposed metrics (See our attached video for more experiments' examples).

\subsection{Private Space} 
\begin{figure}[ht]
\begin{minipage}[ht]{.5\textwidth}
  \vspace*{\fill}
  \centering
  \includegraphics[width=0.74\linewidth]{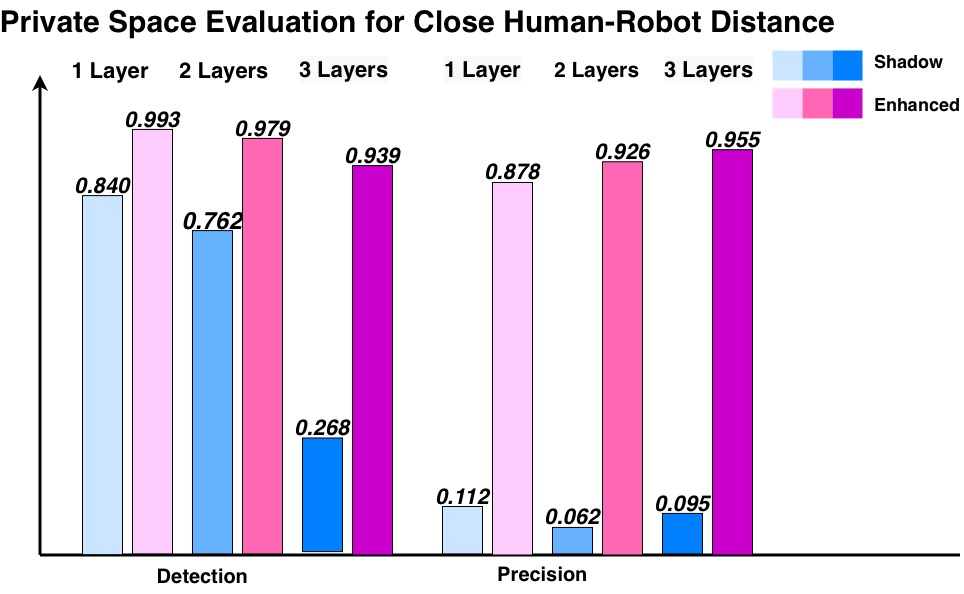}
\end{minipage}
\begin{minipage}[ht]{.5\textwidth}
  \vspace*{\fill}
  \centering
  \includegraphics[width=0.74\linewidth]{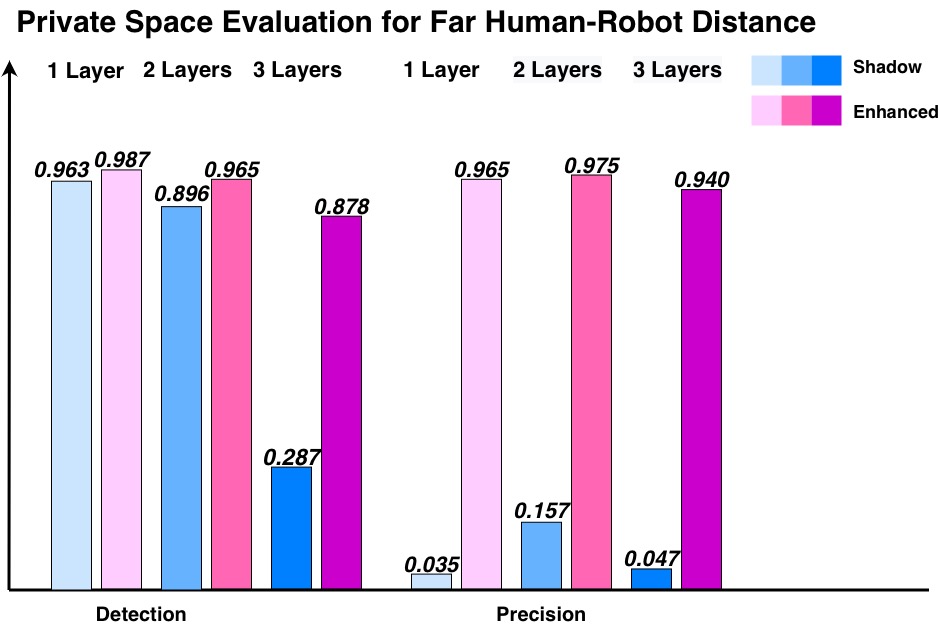}
  \caption{The Evaluation histograms in private spaces. Top to Bottom: Detection Rate and precision for different filter layers on shadow and enhanced images when the distance between the participants and the camera is $2$ and $4$ meters respectively.}
  \label{fig:p_c_l}\par\vfill
\end{minipage}
\end{figure}

\begin{figure}[H]
\begin{minipage}[H]{.5\textwidth}
  \vspace*{\fill}
  \centering
  \includegraphics[width=\linewidth]{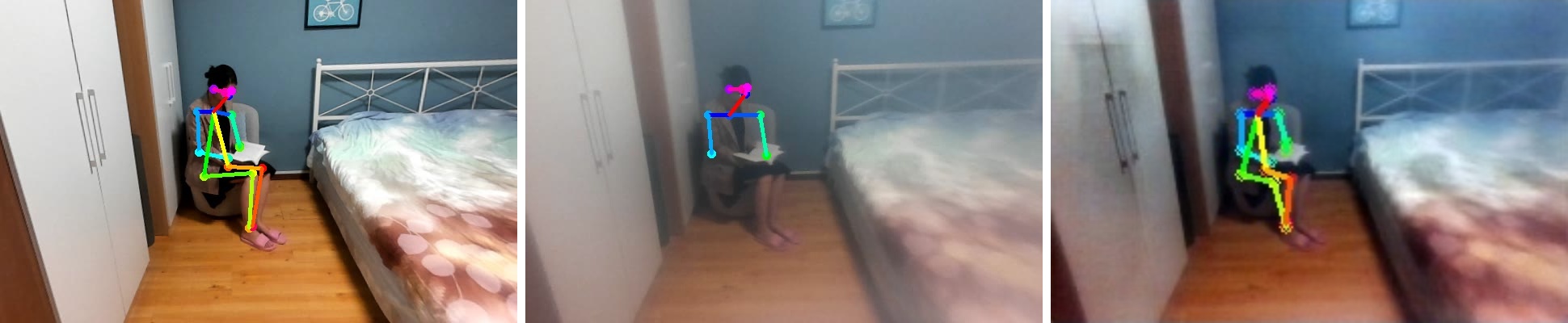}
  \caption{Private space evaluation: the female participant sits $4$ meters away from the camera system covered with $2$ filter layers while reading in her bedroom. The images from left to right show the pose estimation results on the ground-truth clear image, the paired shadow image and enhanced image respectively.}
  \label{fig:p_l}\par\vfill
\end{minipage}
\end{figure}

We evaluate the effectiveness of our method for two different proximity settings and three layer settings.
Figure~\ref{fig:p_c_l} shows the mean DR and SmAP scores for each chosen filter layers on the private space dataset (low background clutter). Comparing scores for the original filtered (Shadow) and the enhanced images, we can see that the detection rate for shadow images degrades with increasing filter layers. SmAP (precision) is very low even with a single filter layer. In comparison, the enhanced images maintain high detection and accuracy scores even with increasing layers. 
Figure~\ref{fig:p_l} shows an example image, comparing the pose estimation on the clear, filtered, and enhanced versions of the same image. 

Besides, from the histograms in Figure~\ref{fig:p_c_l}, we find that proximity does not have a consistent effect on the shadow images' pose estimation results. Both detection rates and precision are similar for people standing closer and farther from the camera. 

\subsection{Public Space}
\begin{figure}[ht]
  \vspace*{\fill}
  \centering
  \includegraphics[width=\linewidth]{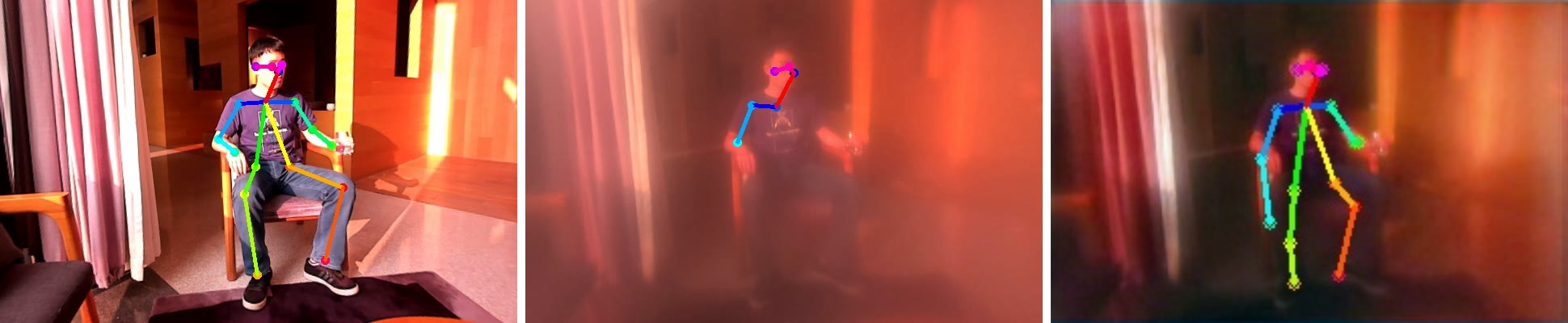}
  \caption{Public space evaluation: the male participant sits $2$ meters away from the camera system covered with $3$ filter layers while drinking in a cafe. The images from left to right show the pose estimation results on the ground-truth clear image, the paired shadow image and enhanced image respectively.}
  \label{fig:u_c}
\end{figure}

Figure~\ref{fig:u_c} shows an example of an image taken in the public (cluttered) environment, demonstrating the pose estimation performance on the clear, filtered, and enhanced versions of the same image. 

\begin{figure}[ht]
\begin{minipage}[ht]{.5\textwidth}
  \vspace*{\fill}
  \centering
  \includegraphics[width=0.74\linewidth]{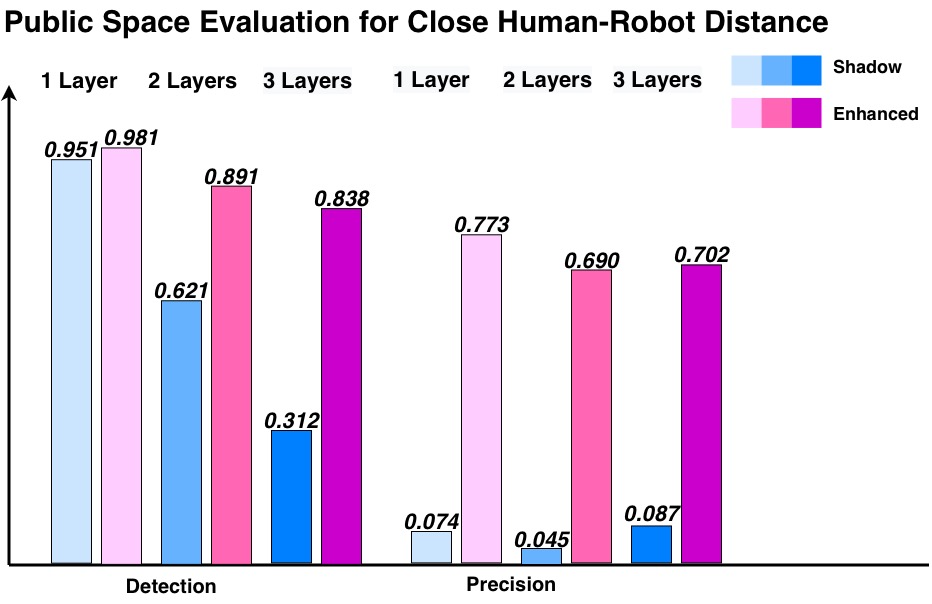}
\end{minipage}
\begin{minipage}[ht]{.5\textwidth}
  \vspace*{\fill}
  \centering
   \captionsetup{margin=0.35cm}
  \includegraphics[width=0.74\linewidth]{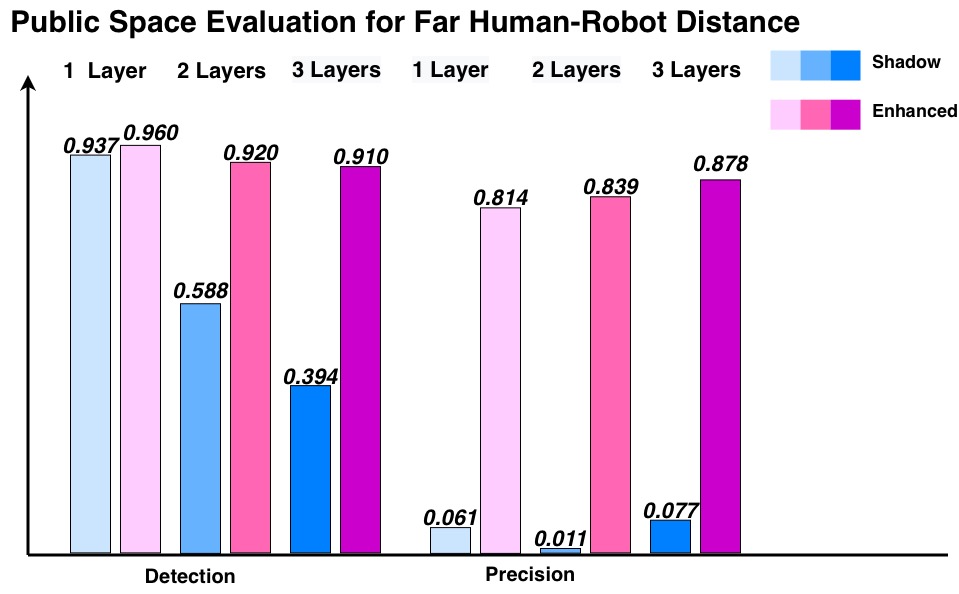}
  \caption{The Evaluation histograms in public spaces. Top to Bottom: Detection Rate and precision for different filter layers on shadow and enhanced images when the distance between the participants and the camera is $2$ and $4$ meters respectively.}
  \label{fig:u_c_l}\par\vfill
\end{minipage}
\end{figure}

The results, seen in Figure~\ref{fig:u_c_l}, indicate a similar trend as in the private spaces. 
That said, un-enhanced shadow images degrade faster in public spaces, with loss of detection rates happening with $2$ layers of filtering. 
In addition, compared with private spaces, for the same filter layers
and proximity, the percentage of precisely detected key points on enhanced images in public spaces is smaller than the percentage in private spaces. Therefore, the public spaces are more challenging for image enhancement than private spaces.

\subsection{Ablation Study}
To evaluate each component's importance in our defined loss function, we conduct an ablation study for our proposed network's loss function. During the ablation study, to make more obvious distinctions of our network's performances between original and modified loss functions, we choose datasets created from $2$ filter layers and
$3$ filter layers. Specifically, for each type of spaces, we choose $3610$ paired images generated from $3$ filter layers and $3956$ paired images generated from $2$ filter layers. Then by changing the optimization objective in the enhancement module previously associated with $\mathcal{L}_{SL}$ to
$\mathcal{L}_{PL}+\mathcal{L}_{EL}$, we remove the structural loss from our optimization objective and retrain the network from scratch. Similarly, we train our network by removing the perceptual loss and edge loss respectively. In the ablation study table we list below, we use $i_{o}$ to represent the original enhanced image output by the proposed network for $i$ filter layers with $i \in \{1,2,3\}$, $i_{g}$, $i_{p}$ and $i_{t}$ to represent the enhanced image we obtain by removing one of the edge loss, perceptual loss, structural loss respectively.
\begin{table}[h]
\huge
\begin{adjustbox}{width=\columnwidth,center}
\begin{tabular}{|c|c|c|c|c|c|c|c|c|}
\hline
$\text{Filler}_{\text{Loss}}$/Metric & $2_{o}$  & $2_{t}$  & $2_{g}$   & $2_{p}$  & $3_{o}$  & $3_{t}$   & $3_{g}$  & $3_{p}$  \\ \hline
DR                         & 0.959 & 0.841 & 0.944  & 0.557 & 0.954 & 0.846  & 0.920 & 0.693 \\ \hline
SmAP                        & 0.901 & 0.748  & 0.895 & 0.633 & 0.887 & 0.758 & 0.838 & 0.597 \\ \hline
\end{tabular}
\end{adjustbox}
\caption{The ablation study table we obtain by removing one of the perceptual loss, structural loss and edge loss.}
\label{table:ablation_study}
\end{table}
\iffalse
\begin{figure}[h]
\centering
\begin{subfigure}{.25\textwidth}
  \centering
  \includegraphics[width=.9\linewidth]{Figs/ab_r_o.png}
  %\caption{The camera system we used.}
\end{subfigure}%
\begin{subfigure}{.25\textwidth}
  \centering
  \includegraphics[width=.9\linewidth]{Figs/ab_r_s.png}
  %\caption{The camera system with plastic bags on it.}
\end{subfigure}
\caption{The left subfigure shows the pose estimation on the enhanced image using the original loss functions for \textit{MiniRes} network optimization . The right subfigure shows pose estimation on the enhanced image using $\mathcal{L}_{PL}+\mathcal{L}_{EL}$ for \textit{MiniRes} network optimization }
\label{fig:ab_r}
\end{figure}
\begin{figure}[h]
\centering
\begin{subfigure}{.25\textwidth}
  \centering
  \includegraphics[width=.9\linewidth]{Figs/ab_d_o.png}
  %\caption{The camera system we used.}
\end{subfigure}%
\begin{subfigure}{.25\textwidth}
  \centering
  \includegraphics[width=.9\linewidth]{Figs/ab_d_p.png}
  %\caption{The camera system with plastic bags on it.}
\end{subfigure}
\caption{The left subfigure shows the pose estimation on the enhanced image using the original loss functions for \textit{MiniDense} network optimization . The right subfigure shows pose estimation on the enhanced image using $\mathcal{L}_{SL}+\mathcal{L}_{EL}$ for \textit{MiniDense} network optimization }
\label{fig:ab_d}
\end{figure}
\fi

We can infer two implications from the ablation study table (Table \ref{table:ablation_study}).
\begin{inparaenum}
\item The perceptual loss is more important than edge loss and structural loss for the performance of our proposed network since $i_{p}$ suffers from the greatest decline in DR and SmAP scores compared with $i_{o}$, with edge loss seemingly being the least contributing factor;  
\item Since scores for the original enhanced images are higher than the enhanced images we obtain by removing structural loss, perceptual loss, and edge loss, respectively, the three components in our loss functions contribute to the network's optimization process.
\end{inparaenum}

\subsection{Comparison to Existing Image Enhancement Methods}
\label{subsec:compare_contrast}
There is no existing method proposed for boosting pose estimation's performances on the type of shadow images we are interested in this work. That said, given the similarity of shadow images to hazy and low-light images in feature space, we evaluate the performances of two state-of-the-art methods focused on enhancing hazy and low-light images.  Specifically, we choose AODNet~\cite{li2017aod}, which focuses on hazy image enhancement, and MBLLEN~\cite{lv2018mbllen}, which focuses on low-light image enhancement for comparison study. Then we use the same paired images as the ablation study. Finally, we evaluate the pose estimation results (\textit{i.e.}, DR, and SmAP scores) of the enhanced images generated from the two state-of-the-art methods using \textit{OpenPose}. 

Table~\ref{table:sota} shows the performance of our method ($i_{e}$) for $i$ layers of filters, $i \in \{1,2,3\}$. $i_{m}$ indicated the output image that MBLLEN produces and $i_{a}$ the output image that AODNet obtains. We can conclude that our proposed network boosts pose estimation on collected shadow images more effectively than AODNet and MBLLEN.
\begin{table}[h]
\large
\begin{adjustbox}{width=\columnwidth,center}
\begin{tabular}{|c|c|c|c|c|c|c|}
\hline
$\text{Layer}_{\text{Methods}}$/Metric & $2_{e}$  & $2_{m}$  & $2_{a}$  & $3_{e}$  & $3_{m}$  & $3_{a}$  \\ \hline
DR                          & 0.959 & 0.605 & 0.730 & 0.954 & 0.531 & 0.582 \\ \hline
SmAP                        & 0.901 & 0.069 & 0.049 & 0.887 & 0.044 & 0.062 \\ \hline
\end{tabular}
\end{adjustbox}
\caption{The evaluation table which shows the performances of AODNet and MBLLEN on the selected shadow image dataset.}
\label{table:sota}
\end{table}
\iffalse
\begin{figure}[h]
\centering
\begin{subfigure}{.25\textwidth}
  \centering
  \includegraphics[width=.9\linewidth]{Figs/sota_d_o.png}
  %\caption{The camera system we used.}
\end{subfigure}%
\begin{subfigure}{.25\textwidth}
  \centering
  \includegraphics[width=.9\linewidth]{Figs/sota_d.png}
  %\caption{The camera system with plastic bags on it.}
\end{subfigure}
\caption{The left subfigure shows the pose estimation on the enhanced image output by MBLLEN network . The right subfigure shows pose estimation on the enhanced image output by \textit{MiniDense} network.}
\label{fig:sota_d}
\end{figure}
\begin{figure}[h]
\centering
\begin{subfigure}{.25\textwidth}
  \centering
  \includegraphics[width=.9\linewidth]{Figs/sota_r_0.JPG}
  %\caption{The camera system we used.}
\end{subfigure}%
\begin{subfigure}{.25\textwidth}
  \centering
  \includegraphics[width=.9\linewidth]{Figs/sota_r.png}
  %\caption{The camera system with plastic bags on it.}
\end{subfigure}
\caption{The left subfigure shows the pose estimation on the enhanced image output by AODNet . The right subfigure shows pose estimation on the enhanced image output by \textit{MiniRes} network.}
\label{fig:sota_r}
\end{figure}
\fi

%%YUHAN TODO: Please offer some design suggestion about filters choices and the suggestions about network architectures here about different scenarios e.g. spaves, proximity and the network structures.
\section{Discussion}
Based on our evaluation results, we offer future researchers potential directions about configuring the privacy-protected camera system.
\begin{itemize}
    \item \textbf{Proximity}: From the four evaluation histograms in Figure~\ref{fig:p_c_l} and~\ref{fig:u_c_l}, we conclude that the system can protect humans' privacy and have robust performances on pose estimation when the distance between humans and camera system is between $2$ and $4$ meters. Therefore, we suggest that future designers can choose the proximity within range $[2,4]$. Also, they are encouraged to explore a more suitable distance between humans and the system to balance the performances on privacy protection and pose estimation for their purposes;
    \item \textbf{Filters}: According to Table~\ref{table:sseq} and the four histograms, we can conclude that $1$ filter layer is more optimal for pose estimation, while $3$ filter layers are more optimal for privacy protection. $2$ filter layers, on the other hand, has the medium score of shadow ratio and can be potentially chosen for balancing between privacy protection and pose estimation.
    Therefore, based on the SR value of Table~\ref{table:sseq}, we propose that designers should choose filters with $\text{SR} \in [0.36,0.45]$ which is the SR value between the $2$ and $3$ filter layers.
\end{itemize}
\section{Conclusion}
This paper proposes a camera system that can protect humans' privacy while obtaining their pose information effectively.
It is based on an image enhancement neural network that uses structural, perceptual, and edge information and is trained on an off-context dataset.
We evaluate the filter layers' privacy protection ability and our proposed network's robustness and find that our method regains the possibility for pose estimation even with a privacy filter. 
This can provide a useful solution for users wanting to cover a robot's camera with a physical filter, but still interact with it. Finally, we offer future researchers potential directions about suitable filter choices and proximity between the camera system and humans. In future, we hope to incorporate social factors into our system design consideration by conducting human studies.

%%%%%%%%%%%%%%%%%%%%%%%%%%%%%%%%%%%%%%%%%%%%%%%%%%%%%%%%%%%%%%%%%%%%%%%%%%%%%%%%

\bibliographystyle{plain}
\bibliography{reference}

\begin{thebibliography}{10}

\bibitem{abdullah2007dynamic}
Mohammad Abdullah-Al-Wadud, Md~Hasanul Kabir, M~Ali~Akber Dewan, and Oksam
  Chae.
\newblock A dynamic histogram equalization for image contrast enhancement.
\newblock {\em IEEE Transactions on Consumer Electronics}, 53(2):593--600,
  2007.

\bibitem{andriluka20142d}
Mykhaylo Andriluka, Leonid Pishchulin, Peter Gehler, and Bernt Schiele.
\newblock 2d human pose estimation: New benchmark and state of the art
  analysis.
\newblock In {\em Proceedings of the IEEE Conference on computer Vision and
  Pattern Recognition}, pages 3686--3693, 2014.

\bibitem{caine2012effect}
Kelly Caine, Selma {\v{S}}abanovic, and Mary Carter.
\newblock The effect of monitoring by cameras and robots on the privacy
  enhancing behaviors of older adults.
\newblock In {\em Proceedings of the seventh annual ACM/IEEE international
  conference on Human-Robot Interaction}, pages 343--350, 2012.

\bibitem{cao2019openpose}
Zhe Cao, Gines~Hidalgo Martinez, Tomas Simon, Shih-En Wei, and Yaser~A Sheikh.
\newblock Openpose: Realtime multi-person 2d pose estimation using part
  affinity fields.
\newblock {\em IEEE Transactions on Pattern Analysis and Machine Intelligence},
  2019.

\bibitem{celik2011contextual}
Turgay Celik and Tardi Tjahjadi.
\newblock Contextual and variational contrast enhancement.
\newblock {\em IEEE Transactions on Image Processing}, 20(12):3431--3441, 2011.

\bibitem{cheng2004simple}
HD~Cheng and XJ~Shi.
\newblock A simple and effective histogram equalization approach to image
  enhancement.
\newblock {\em Digital signal processing}, 14(2):158--170, 2004.

\bibitem{deng2009imagenet}
Jia Deng, Wei Dong, Richard Socher, Li-Jia Li, Kai Li, and Li~Fei-Fei.
\newblock Imagenet: A large-scale hierarchical image database.
\newblock In {\em 2009 IEEE conference on computer vision and pattern
  recognition}, pages 248--255. Ieee, 2009.

\bibitem{fan2005novel}
Jianping Fan, Hangzai Luo, Mohand-Said Hacid, and Elisa Bertino.
\newblock A novel approach for privacy-preserving video sharing.
\newblock In {\em Proceedings of the 14th ACM international conference on
  Information and knowledge management}, pages 609--616, 2005.

\bibitem{fu2005fast}
Haoying Fu, Michael~K Ng, Mila Nikolova, Jesse Barlow, and Wai-ki Ching.
\newblock Fast algorithms for l1 norm/mixed l1 and l2 norms for image
  restoration.
\newblock In {\em International Conference on Computational Science and Its
  Applications}, pages 843--851. Springer, 2005.

\bibitem{fu2016fusion}
Xueyang Fu, Delu Zeng, Yue Huang, Yinghao Liao, Xinghao Ding, and John Paisley.
\newblock A fusion-based enhancing method for weakly illuminated images.
\newblock {\em Signal Processing}, 129:82--96, 2016.

\bibitem{fu2016weighted}
Xueyang Fu, Delu Zeng, Yue Huang, Xiao-Ping Zhang, and Xinghao Ding.
\newblock A weighted variational model for simultaneous reflectance and
  illumination estimation.
\newblock In {\em Proceedings of the IEEE Conference on Computer Vision and
  Pattern Recognition}, pages 2782--2790, 2016.

\bibitem{goodrich2008human}
Michael~A Goodrich and Alan~C Schultz.
\newblock {\em Human-robot interaction: a survey}.
\newblock Now Publishers Inc, 2008.

\bibitem{he2016deep}
Kaiming He, Xiangyu Zhang, Shaoqing Ren, and Jian Sun.
\newblock Deep residual learning for image recognition.
\newblock In {\em Proceedings of the IEEE conference on computer vision and
  pattern recognition}, pages 770--778, 2016.

\bibitem{herath2017going}
Samitha Herath, Mehrtash Harandi, and Fatih Porikli.
\newblock Going deeper into action recognition: A survey.
\newblock {\em Image and vision computing}, 60:4--21, 2017.

\bibitem{shadowsense}
Yuhan Hu, Sara Bejarano, and Guy Hoffman.
\newblock Shadowsense: Detecting human touch in a social robot through shadow
  image processing.
\newblock {\em Proceedings of the ACM on Interactive, Mobile, Wearable and
  Ubiquitous Technologies}, 1(1), 2020.

\bibitem{kohli2013key}
Pushmeet Kohli and Jamie Shotton.
\newblock Key developments in human pose estimation for kinect.
\newblock In {\em consumer depth cameras for computer vision}, pages 63--70.
  Springer, 2013.

\bibitem{li2017aod}
Boyi Li, Xiulian Peng, Zhangyang Wang, Jizheng Xu, and Dan Feng.
\newblock {AOD}-net: All-in-one dehazing network.
\newblock In {\em Proceedings of the IEEE international conference on computer
  vision}, pages 4770--4778, 2017.

\bibitem{li2018benchmarking}
Boyi Li, Wenqi Ren, Dengpan Fu, Dacheng Tao, Dan Feng, Wenjun Zeng, and
  Zhangyang Wang.
\newblock Benchmarking single-image dehazing and beyond.
\newblock {\em IEEE Transactions on Image Processing}, 28(1):492--505, 2018.

\bibitem{li2020underwater}
Chongyi Li, Saeed Anwar, and Fatih Porikli.
\newblock Underwater scene prior inspired deep underwater image and video
  enhancement.
\newblock {\em Pattern Recognition}, 98:107038, 2020.

\bibitem{li2019heavy}
Ruoteng Li, Loong-Fah Cheong, and Robby~T Tan.
\newblock Heavy rain image restoration: Integrating physics model and
  conditional adversarial learning.
\newblock In {\em Proceedings of the IEEE Conference on Computer Vision and
  Pattern Recognition}, pages 1633--1642, 2019.

\bibitem{liu2014no}
Lixiong Liu, Bao Liu, Hua Huang, and Alan~Conrad Bovik.
\newblock No-reference image quality assessment based on spatial and spectral
  entropies.
\newblock {\em Signal Processing: Image Communication}, 29(8):856--863, 2014.

\bibitem{lv2018mbllen}
Feifan Lv, Feng Lu, Jianhua Wu, and Chongsoon Lim.
\newblock Mbllen: Low-light image/video enhancement using cnns.
\newblock In {\em BMVC}, page 220, 2018.

\bibitem{lyu2017privacy}
Lingjuan Lyu, Xuanli He, Yee~Wei Law, and Marimuthu Palaniswami.
\newblock Privacy-preserving collaborative deep learning with application to
  human activity recognition.
\newblock In {\em Proceedings of the 2017 ACM on Conference on Information and
  Knowledge Management}, pages 1219--1228, 2017.

\bibitem{neustaedter2006blur}
Carman Neustaedter, Saul Greenberg, and Michael Boyle.
\newblock Blur filtration fails to preserve privacy for home-based video
  conferencing.
\newblock {\em ACM Transactions on Computer-Human Interaction (TOCHI)},
  13(1):1--36, 2006.

\bibitem{pisano1998contrast}
Etta~D Pisano, Shuquan Zong, Bradley~M Hemminger, Marla DeLuca, R~Eugene
  Johnston, Keith Muller, M~Patricia Braeuning, and Stephen~M Pizer.
\newblock Contrast limited adaptive histogram equalization image processing to
  improve the detection of simulated spiculations in dense mammograms.
\newblock {\em Journal of Digital imaging}, 11(4):193, 1998.

\bibitem{qian1997binocular}
Ning Qian.
\newblock Binocular disparity and the perception of depth.
\newblock {\em Neuron}, 18(3):359--368, 1997.

\bibitem{Ren2016}
Wenqi Ren, Si~Liu, Hua Zhang, Jinshan Pan, Xiaochun Cao, and Ming-Hsuan Yang.
\newblock Single image dehazing via multi-scale convolutional neural networks.
\newblock In {\em Computer Vision {\textendash} {ECCV} 2016}, pages 154--169.
  Springer International Publishing, 2016.

\bibitem{ren2020single}
Wenqi Ren, Jinshan Pan, Hua Zhang, Xiaochun Cao, and Ming-Hsuan Yang.
\newblock Single image dehazing via multi-scale convolutional neural networks
  with holistic edges.
\newblock {\em International Journal of Computer Vision}, 128(1):240--259,
  2020.

\bibitem{sarafianos20163d}
Nikolaos Sarafianos, Bogdan Boteanu, Bogdan Ionescu, and Ioannis~A Kakadiaris.
\newblock 3d human pose estimation: A review of the literature and analysis of
  covariates.
\newblock {\em Computer Vision and Image Understanding}, 152:1--20, 2016.

\bibitem{schalkoff1989digital}
Robert~J Schalkoff.
\newblock {\em Digital image processing and computer vision}, volume 286.
\newblock Wiley New York, 1989.

\bibitem{shakhnarovich2003fast}
Gregory Shakhnarovich, Paul Viola, and Trevor Darrell.
\newblock Fast pose estimation with parameter-sensitive hashing.
\newblock In {\em null}, page 750. IEEE, 2003.

\bibitem{srivastav2019human}
Vinkle Srivastav, Afshin Gangi, and Nicolas Padoy.
\newblock Human pose estimation on privacy-preserving low-resolution depth
  images.
\newblock In {\em International Conference on Medical Image Computing and
  Computer-Assisted Intervention}, pages 583--591. Springer, 2019.

\bibitem{wang2013naturalness}
Shuhang Wang, Jin Zheng, Hai-Miao Hu, and Bo~Li.
\newblock Naturalness preserved enhancement algorithm for non-uniform
  illumination images.
\newblock {\em IEEE Transactions on Image Processing}, 22(9):3538--3548, 2013.

\bibitem{wang2004image}
Zhou Wang, Alan~C Bovik, Hamid~R Sheikh, and Eero~P Simoncelli.
\newblock Image quality assessment: from error visibility to structural
  similarity.
\newblock {\em IEEE transactions on image processing}, 13(4):600--612, 2004.

\end{thebibliography}
\end{document}